\documentclass[conference]{IEEEtran}
\IEEEoverridecommandlockouts

\usepackage{cite}
\usepackage{amsmath,amssymb,amsfonts}
\usepackage{algorithmic}
\usepackage{graphicx}
\usepackage{textcomp}
\usepackage{mhchem}
\usepackage{xcolor}
\usepackage{caption}
\usepackage{bm}

\usepackage[T1]{fontenc}
\def\BibTeX{{\rm B\kern-.05em{\sc i\kern-.025em b}\kern-.08em
    T\kern-.1667em\lower.7ex\hbox{E}\kern-.125emX}}
\begin{document}

\title{Model Predictive Control Design of a 3-DOF Robot Arm Based on Recognition of Spatial Coordinates \\
}

\author{\IEEEauthorblockN{Zhangxi Zhou}
\IEEEauthorblockA{\textit{Dept. of Electrical Eng \& Electronics}\\ 
\textit{University of Liverpool}\\
Liverpool, United Kingdom \\
https://orcid.org/0000-0002-1529-3710}
\and
\IEEEauthorblockN{Yuyao Zhang}
\IEEEauthorblockA{\textit{Dept. of Electrical Eng \& Electronics} \\
\textit{University of Liverpool}\\
Liverpool, United Kingdom \\
https://orcid.org/0000-0001-6444-4215}
\and
\IEEEauthorblockN{Yezhang Li}
\IEEEauthorblockA{\textit{Dept. of Mathematical Sciences} \\
\textit{University of Liverpool}\\
Liverpool, United Kingdom \\
https://orcid.org/0000-0002-5740-5477}
}

\maketitle

\begin{abstract}
This paper uses Model Predictive Control (MPC) to optimise the input torques of a Three-Degrees-of-Freedom (DOF) robotic arm, enabling it to operate to the target position and grasp the object accurately. A monocular camera is firstly used to recognise the colour and depth of the object. Then, the inverse kinematics calculation and the spatial coordinates of the object through coordinate transformation are combined to get the required rotating angle of each servo. Finally, the dynamic model of the robotic arm structure is derived and the model predictive control is applied to simulate the optimal input torques of servos to minimize the cost function.
\end{abstract}


\section{Introduction}
Robotic arms can be widely used in various areas, such as industry, aerospace and medical care. During the last few decades, the control and design of robotic arms have become major fields of interest within the field of robotics. The technologies of robotic arms consist of computer vision, trajectory planning, mechanical design, electronic control, and control theories.

In the 21st century, researchers began to focus on algorithms of robotic arms, especially the optimisation of the robotic system. In 2004, Babazadeh and Sadati \cite{b1} proposed an optimisation algorithm for controlling the multiple-arm robotic system using the gradient method, which is suitable for complicated multi-arm systems. As a consequence of the research and development of robotic arms, classical control theory has been developed and applied to robotic arms. In 2015, Akyürek \cite{b2} applied PID control and force feedback to accomplish ambidextrous hand grasping, improving the accuracy and decreasing the iterations. Nowadays, Model Predictive Control (MPC) has been developed to be applied in robotics. As one of the advanced control theories, MPC can be used to optimise the performance of robotic arms. In \cite{b3}, the MPC approach was used to control a three-degrees-of-freedom (DOF) manipulator robot under the consideration of a second-order closed loop system. Reference \cite{b4} applied MPC algorithm and non-industrial robotic arms to develop the high-precision trajectory tracking. In \cite{b5} and \cite{b6}, MPC was used to demonstrate and test the simulation of a 5-DOF robot arm and an autonomous mobile grasping system respectively.

The initial purpose of this project is to use the robotic arm to grasp the object based on the recognised coordinates and calculated joint angle. Firstly, A monocular camera is used for recognising the spatial coordinates of colour-recognised objects. Then, the spatial coordinates are transformed into joint angle output by inverse kinematics calculation. However, in the real-time implementation, due to the disturbance of external forces and the influence of gravity and other factors, the manipulator cannot accurately run to the target position to grasp. In order to improve the stability of the manipulator and make the servo mechanism accurately work to the specified position, the dynamic model of the manipulator is established in this paper. The optimal input torque of the servo rotation is obtained by model predictive control, which makes the grasping process more accurate and stable. Also, the minimisation of the cost function reduces energy consumption.

The article is organized as follows: Section\,II introduces the principle of colour and depth recognition using a monocular camera. In Section\,III, the inverse kinematics model of a 3-DOF robot is presented. Then, combining the above technologies, the scene of automatic object recognition and grasping by a fixed camera is demonstrated. To obtain the state-space model, Section\,IV provides the dynamic model. In Section\,V, MPC is applied to minimise the cost function by using the obtained dynamic model. Finally, the results of the MPC simulation along with the operating process are presented in Section\,VI.

\begin{figure}[htbp]
\centering
\includegraphics[scale=0.55]{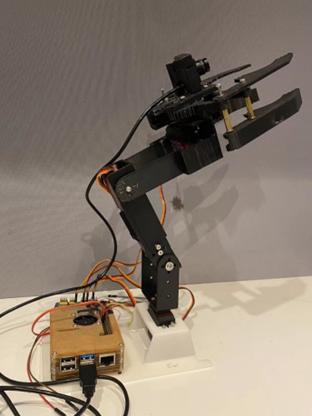}       
\caption{The overall view of the designing robotic arm.}
\label{fig1}
\end{figure}

\section{Computer vision}

\subsection{Principle of colour recognition}
\begin{figure}[htbp]
\centering
\includegraphics[scale=0.67]{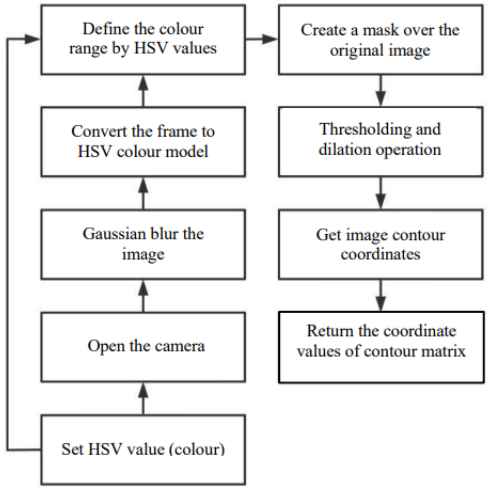}       
\caption{The procedure of the colour recognition.}
\label{fig2}
\end{figure}

Fig.~\ref{fig2} presents the procedure of colour recognition. First, turn on the camera and blur the image initially to facilitate subsequent processing. Colours are then identified and extracted based on HSV values. Then thresholding and dilation are performed to make the image clear. Since the lines are extracted and located, erosion may cause more spaced breakpoints to cut the lines. Therefore, only dilation operation is performed without erosion. Finally, the contour frame is made according to the colour.

\subsection{Principle of depth recognition}
\begin{figure}[htbp]
\centering
\includegraphics[scale=0.35]{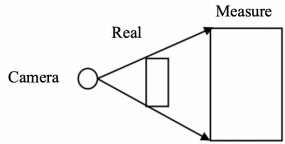}       
\caption{Principle of the depth recognition.}
\label{fig3}
\end{figure}

Fig.~\ref{fig3} shows the principle of depth recognition by a monocular camera. In order to recognise the depth of the object based on colour recognition, the actual width of the object, measured distance and measured pixel width should be known before recognition to calculate the camera focal length first. Camera focal length can be written as:
\begin{equation}
F = \frac{PD}{W} \label{eq}
\end{equation}
\noindent
where $D$ is the measured distance away from the camera, $W$ is
the actual width of the object and $P$ is the measured pixel width.

As the camera continues to move closer or farther away from the target, the real distance of the object from the camera can be calculated by using the similarity transformation:
\begin{equation}
D' = \frac{W'F}{P} \label{eq}
\end{equation}

\vspace{-10pt}

\vspace{-10pt}

\section{Inverse kinematics}
\vspace{-10pt}
\subsection{Denavit-Hartenberg model}
\vspace{-7pt}
The purpose of conducting inverse kinematics calculation is
to transform the spatial coordinates input to the joint angle
output, so that the robotic arm could move to the given spatial
coordinate to grasp the objects. The procedure can be described
as follows. Firstly, the Denavit-Hartenberg (DH) model is used
to establish the structure of the robotic arm. The structure of the
robotic arm refers to \cite{b8}, as shown in Fig.~\ref{fig4}. The links are
labelled as $1$, $2$, $3$, $4$. The servos at links $1$, $2$ and $3$ are
responsible for moving while the servo at link $4$ is used for
grasping. Table.~\ref{tab1} shows the parameters of each link, where $a_i$ and $\alpha_i$ represent the distance and angle of rotation from $\hat{z}_i$ to $\hat{z}_{i+1}$ along the $\hat{x}_i$-axis, and $d_i$ and $\theta_i(q_i)$ represent the distance and angle of rotation from $\hat{x}_{i-1}$ to $\hat{x}_{i}$ along the $\hat{z}_i$-axis, respectively.

As \cite{b7} proposed, the transformation matrix of coordinate system ${i}$ relative to ${i-1}$ is

\begin{equation}
\ce{^{i-1}_{i}}T = 
\left[\begin{array}{@{}cccc@{}}
  \begin{matrix}
  c\theta_i \\
  s\theta_i c\alpha_{i-1} \\
  s\theta_i s\alpha_{i-1} \\
  0
  \end{matrix}
  & 
  \begin{matrix}
  -s\theta_i \\
  c\theta_i c\alpha_{i-1} \\
  c\theta_i s\alpha_{i-1} \\
  0
  \end{matrix}
  & 
  \begin{matrix}
  0 \\
  -s\alpha_{i-1} \\
  c\alpha_{i-1} \\
  0
  \end{matrix}
  & 
  \begin{matrix}
  a_{i-1} \\
  -s\alpha_{i-1}d_i \\
  c\alpha_{i-1}d_i \\
  1
  \end{matrix}
\end{array}\right]\label{eq}
\end{equation}

In the following section, $\cos{\theta_i}$ and $\sin{\theta_i}$ are represented as $c\theta_i$ and $s\theta_i$ (or $c_i$ and $s_i$) for simplification. Similarly, $c_{ij}$ and $s_{ij}$ are used for representing $\cos{(\theta_i + \theta_j)}$ and $\sin{(\theta_i + \theta_j)}$ respectively. By substituting the parameters in Table.~\ref{tab1} into (3), $\ce{^{i-1}_{i}}T$ can be obtained as:

\vspace{10pt}

\begin{figure}[htbp]
\centering
\includegraphics[scale=0.52]{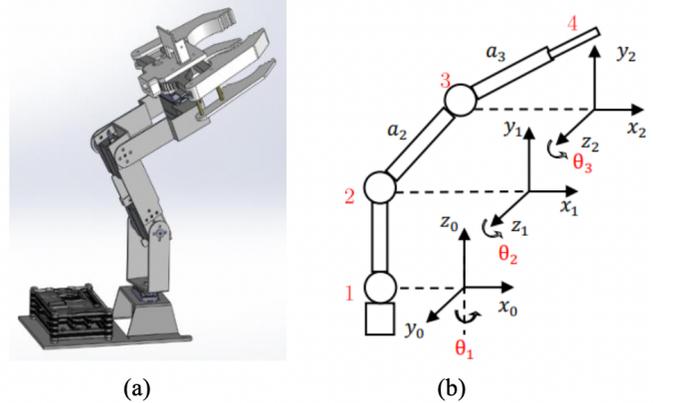}       
\caption{Structure of the 3-DOF robotic arm.}
\label{fig4}
\end{figure}

\begin{table}[htbp]
\caption{The parameters of each link}
\begin{center}

\setlength{\tabcolsep}{5mm}{
\begin{tabular}{|c|c|c|c|c|}
\hline
\textbf{Links}&\multicolumn{4}{|c|}{\text{ }\textbf{Parameters}\text{ }} \\
\cline{2-5} 
\textbf{Number} & \text{ }\bm{$a_i$}\text{ } & \text{ }\bm{$\alpha_i$}\text{ } & \text{ }\bm{$d_i$}\text{ } & \text{ }\bm{$\theta_i(q_i)$}\text{ } \\
\hline
   1    & 0 & $\pi/2$ & 0 & $\theta_1(q_1)$ \\
\hline
2 & $a_2$ & 0 & 0 & $\theta_2(q_2)$ \\
\hline
3 & $a_3$ & 0 & 0 & $\theta_3(q_3)$  \\
\hline
4 & 0 & 0 & 0 & 0  \\
\hline

\end{tabular}}
\label{tab1}
\end{center}
\end{table}

\begin{equation*}
_1^{0}T = 
\left[\begin{array}{@{}cccc@{}}
  \begin{matrix}
  c\theta_1 \\
  s\theta_1 \\
  0 \\
  0
  \end{matrix}
  & 
  \begin{matrix}
  -s\theta_1 \\
  c\theta_1 \\
  0 \\
  0
  \end{matrix}
  & 
  \begin{matrix}
  0 \\
  0 \\
  1 \\
  0
  \end{matrix}
  & 
  \begin{matrix}
  0 \\
  0 \\
  0 \\
  1
  \end{matrix}
\end{array}\right], \text{}
_2^{1}T = 
\left[\begin{array}{@{}cccc@{}}
  \begin{matrix}
  c\theta_2 \\
  0 \\
  s\theta_2 \\
  0
  \end{matrix}
  & 
  \begin{matrix}
  -s\theta_2 \\
  0 \\
  c\theta_2 \\
  0
  \end{matrix}
  & 
  \begin{matrix}
  0 \\
  -1 \\
  0 \\
  0
  \end{matrix}
  & 
  \begin{matrix}
  0 \\
  0 \\
  0 \\
  1
  \end{matrix}
\end{array}\right]
\end{equation*}
\begin{equation}
_3^{2}T = 
\left[\begin{array}{@{}cccc@{}}
  \begin{matrix}
  c\theta_3 \\
  s\theta_3 \\
  0 \\
  0
  \end{matrix}
  & 
  \begin{matrix}
  -s\theta_3 \\
  c\theta_3 \\
  0 \\
  0
  \end{matrix}
  & 
  \begin{matrix}
  0 \\
  0 \\
  1 \\
  0
  \end{matrix}
  & 
  \begin{matrix}
  a_2 \\
  0 \\
  0 \\
  1
  \end{matrix}
\end{array}\right], \text{}
_4^{3}T = 
\left[\begin{array}{@{}cccc@{}}
  \begin{matrix}
  1 \\
  0 \\
  0 \\
  0
  \end{matrix}
  & 
  \begin{matrix}
  0 \\
  1 \\
  0 \\
  0
  \end{matrix}
  & 
  \begin{matrix}
  0 \\
  0 \\
  1 \\
  0
  \end{matrix}
  & 
  \begin{matrix}
 a_3 \\
  0 \\
  0 \\
  1
  \end{matrix}
\end{array}\right]\label{eq}
\end{equation}
After the recursion, the matrix $_4^0T$ can be derived as:

\begin{equation}
_4^{0}T = _1^{0}T _2^{1}T _3^{2}T _4^{3}T = 
\left[\begin{array}{@{}cccc@{}}
  \begin{matrix}
  r_{11} \\
  r_{21} \\
  r_{31} \\
  0
  \end{matrix}
  & 
  \begin{matrix}
  r_{12} \\
  r_{22} \\
  r_{32} \\
  0
  \end{matrix}
  & 
  \begin{matrix}
  r_{13} \\
  r_{23} \\
  r_{33} \\
  0
  \end{matrix}
  & 
  \begin{matrix}
  p_x \\
  p_y \\
  p_z \\
  1
  \end{matrix}
\end{array}\right]\label{eq}
\end{equation}
The elements of $_4^0T$ hold the physical meaning: $r_{ij}$ ($i, j = 1, 2, 3$) represent the rotation matrix, and $p_x$, $p_y$, $p_z$ represent the spatial coordinates. The purpose of this calculation is to use $p_x$, $p_y$, $p_z$ to represent $\theta_1$, $\theta_2$ and $\theta_3$.
The rotation matrix $r_{ij}$, ($i, j = 1, 2, 3$) is:
\begin{equation}
\left[\begin{array}{@{}ccc@{}}
  \begin{matrix}
  c_1c_2c_3-c_1s_2s_3 \\
  c_2s_1c_3-s_2s_1s_3 \\
  s_2c_3+c_2s_3 
  \end{matrix}
  & 
  \begin{matrix}
  -c_1c_2s_3-c_1s_2c_3 \\
  -c_2s_1s_3-s_2s_1c_3 \\
  -s_2s_3+c_2c_3 
  \end{matrix}
  & 
  \begin{matrix}
  0 \\
  -c_1 \\
  0
  \end{matrix}
\end{array}\right]\label{eq}
\end{equation}
\noindent
and the spatial coordinates are:
\begin{subequations}\label{eq:litdiff}
\begin{align}
p_x &= a_3(c_1c_2c_3 - c_1s_2s_3)+c_1c_2a_2\label{eq} \\
p_y &= a_3(c_2s_1c_3 - s_2s_1s_3)+c_2s_1a_2\label{eq} \\
p_z &= a_3s_{23}+s_2a_2\label{eq}
\end{align}
\end{subequations}
By summing squares of (7a), (7b) and (7c):
\begin{equation}
p_x^2 + p_y^2 + p_z^2 = a_2^2 + a_3^2 + 2a_2a_3c_3\label{eq}
\end{equation}
\begin{equation}
\Rightarrow c_3 = \frac{p_x^2 + p_y^2 + p_z^2 - \alpha_2^2 - \alpha_3^2}{2a_2a_3}, \text{ } s_3 = \sqrt{1 - c_3^2}\label{eq}
\end{equation}
Therefore,
\begin{equation}
\theta_3 = \arctan{\left(\frac{\sqrt{1 - c_3^2}}{c_3}\right)}\label{eq}
\end{equation}
Since different signs of $\theta$ lead to different solutions, one combination of solutions is chosen here. According to the projection of the end effector of the robot arm on the X-Y plane, the $\theta_1$ can be calculated,
\begin{equation}
\theta_1 = \arctan{\left(\frac{p_x}{p_y}\right)}\label{eq}
\end{equation}
$\theta_2$ can be calculated via the sum of the squares of $p_x$ and $p_y$:
\begin{equation}
p_x^2 + p_y^2 = (a_2c_2 + a_3c_{23})^2\label{eq}
\end{equation}
\begin{equation}
\Rightarrow c_2 = \frac{\sqrt{p_x^2 + p_y^2} + a_3s_2s_3}{a_2 + a_3c_3}\label{eq}
\end{equation}
Then substitute $s_2 = (p_z - a_3c_2s_3)/(a_2+a_3c_3)$ obtained from (7c) into $c_2$:
\begin{equation}
c_2 = \frac{\sqrt{p_x^2 + p_y^2}(a_2 + a_3c_3) + p_za_3s_3}{a_2^2 + a_3^2 + 2a_2a_3c_3}\label{eq}
\end{equation}
Then $\theta_2$ could then be calculated:
\begin{equation}
\theta_2 = \arctan{\left(\frac{-\sqrt{p_x^2 + p_y^2}(a_3s_3) + p_z(a_2 + a_3c_3)}{\sqrt{p_x^2 + p_y^2}(a_2 + a_3c_3) + p_za_3s_3}\right)}\label{eq}
\end{equation}
Therefore, the set of solutions chosen here is shown in (11), (15) and (10). Parameters $c_3$ and $s_3$ can be referred to (9). Part of the derivation process refers to \cite{b8}.
\vspace{-5pt}
\subsection{Combination of computer vision and inverse kinematics}\label{AA}

\begin{figure}[htbp]
\centering
\includegraphics[scale=0.8]{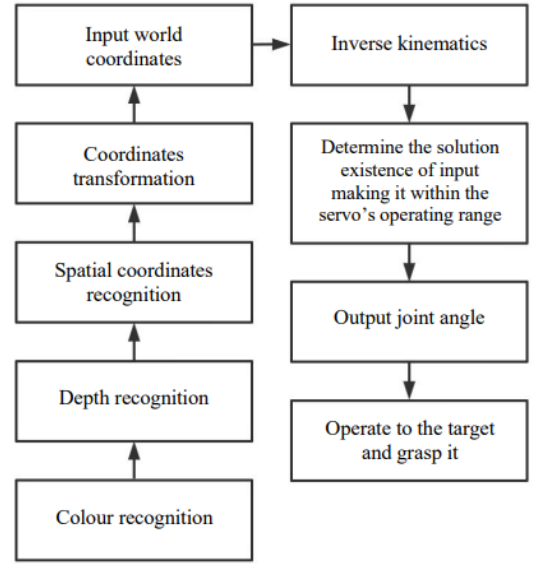}       
\caption{Automatically identify and grasp objects.}
\label{fig5}
\end{figure}

\begin{figure}[htbp]
\centering
\includegraphics[scale=0.25]{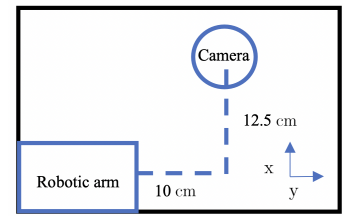}       
\caption{Location plan of the camera and robotic arm. }
\label{fig6}
\end{figure}

 Fig.~\ref{fig5} shows the procedure of automatically identifying objects and grasping them with the fixed camera. Firstly, the spatial coordinates of objects are obtained by colour and depth recognition, and then the camera coordinate system is transformed into the world coordinate system by matrix transformation. Through inverse kinematics calculation, the robotic arm can operate the specific angle to move to the position of the object to grasp. Also, considering the robustness, it is supposed to determine whether the input has a solution and is within the operating range of servos. Fig.~\ref{fig6} presents the location plan of the camera and robotic arm.
\vspace{-10pt}

\vspace{-20pt}

\vspace{220pt}
The height of the camera is $8$\,cm, which means the fixed coordinate of the camera is $(10, 12.5, 8)$\,cm. The position offset of the camera relative to the robotic arm will be used in the matrix transformation.

In the upper left corner of Fig.~\ref{fig7}, a GUI interface that includes six quantities presents the HSV values, which can be adjusted to identify different colours. In the test shown in Fig.~\ref{fig7}, green was chosen as the colour to be identified. The code for colour recognition refers to Heywood \cite{b9}.

The blue frame in Fig.~\ref{fig7} represents the result of colour recognition and the red frame represents the depth recognition. It can be seen that the red and blue frames nearly overlap. This is because the recognition of depth is based on the result of colour recognition. The coordinates in Fig.~\ref{fig7} represent the x-axis, y-axis coordinates and depth respectively. Fig.~\ref{fig8} shows the scene of grasping the target. Within the range of inverse kinematics solutions, the robotic arm can accurately grasp objects.

\section{Dynamic model}
\subsection{State-space model}

According to \cite{b7}, the Lagrangian function of the manipulator can be written as:
\begin{equation}
L = T(q, \dot{q}) - V(q)\label{eq}
\end{equation}
where $T$ is the kinetic energy and $V$ is the potential energy.
\\
Consider the Lagrange equation:
\begin{equation}
\frac{d}{dt}(\frac{\partial L}{\partial \dot{q}}) - (\frac{\partial L}{\partial q}) = \tau \label{eq}
\end{equation}
The standard manipulator dynamic equation is derived as:
\begin{equation}
M(q, \dot{q})\ddot{q} + N(q, \dot{q})\dot{q} + g(q) = \tau \label{eq}
\end{equation}
where $M(q, \dot{q})$ is the inertia matrix, $N(q, \dot{q})$ is the Centrifugal and Coriolis matrix and $g(q)$ is the gravitational vector:
\begin{equation}
\begin{split}
M(q, \dot{q}) &= 
\left[\begin{array}{@{}ccc@{}}
  \begin{matrix}
  d_{11} \\
  d_{21} \\
  d_{31} 
  \end{matrix}
  & 
  \begin{matrix}
  d_{12} \\
  d_{22} \\
  d_{32} 
  \end{matrix}
  & 
  \begin{matrix}
  d_{13} \\
  d_{23} \\
  d_{33} 
  \end{matrix}
\end{array}\right] \\
N(q, \dot{q}) &= [c_1 \text{ } c_2 \text{ } c_3]^T \\
g(q) &= 
\left[\begin{array}{@{}c@{}}
  \begin{matrix}
  0 \\
  g_0
  \left\{\begin{array}{@{}c@{}}
  \begin{matrix}
  m_3(a_2c_2 + (a_3 + x_{c_3})c_{23} - y_{c_3}s_{23}) \\
   + m_2((a_2 + x_{c_2})c_{2} - y_{c_2}s_{2})
  \end{matrix}
  \end{array}\right\} \\
  m_3 g_0\left( (a_3 + x_{c_3})c_{23} - y_{c_3}s_{23} \right) 
  \end{matrix}
\end{array}\right]
\end{split}
\end{equation}
where $g_0 = 9.8\,m\cdot s^{-2}$ and $(x_{c_i}, y_{c_i}, z_{c_i})$ represents the centroid coordinates of link $i$. Due to the space limitations, $M(q, \dot{q})$ and $N(q, \dot{q})$ cannot be shown here. They are related to the inertia tensor of three links:
\begin{equation}
I = 
\left[\begin{array}{@{}ccc@{}}
  \begin{matrix}
  I_{xx} \\
  I_{yx} \\
  I_{zx} 
  \end{matrix}
  & 
  \begin{matrix}
  I_{xy} \\
  I_{yy} \\
  I_{zy} 
  \end{matrix}
  & 
  \begin{matrix}
  I_{xz} \\
  I_{yz} \\
  I_{zz} 
  \end{matrix}
\end{array}\right]
\end{equation}
\\
Rewrite the (19) into the following form:
\begin{equation}
\ddot{q} = -M^{-1}N\dot{q} - M^{-1}g + M^{-1}\tau \label{eq}
\end{equation}

\begin{figure}[htbp]
\centering
\includegraphics[scale=0.35]{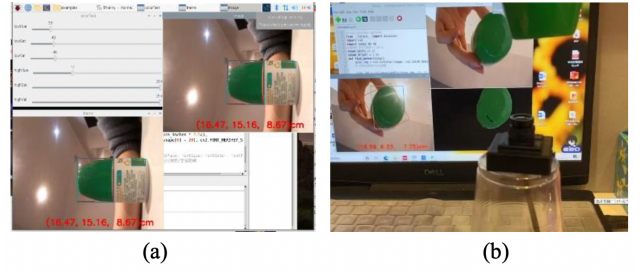}       
\caption{Combination of colour and depth recognition.}
\label{fig7}
\end{figure}
\begin{figure}[htbp]
\centering
\includegraphics[scale=0.3]{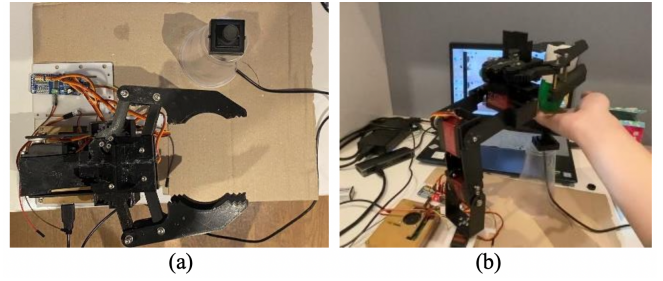}       
\caption{Scene of location and grasping with the camera fixed.}
\label{fig8}
\end{figure}
\noindent
According to \cite{b5}, the controller can be chosen as follows to compose the linear-like state-space model:
\begin{equation}
u = -M^{-1}(\tau - g) \label{eq}
\end{equation}
Then, the state-space model can be derived as:
\begin{equation}
\left[\begin{array}{@{}c@{}}
  \begin{matrix}
  \dot{q} \\
  \ddot{q}
  \end{matrix}
\end{array}\right] = 
\left[\begin{array}{@{}cc@{}}
  \begin{matrix}
  0_{3,3} \\
  0_{3,3}
  \end{matrix}
  & 
  \begin{matrix}
  I_3 \\
  -M^{-1}N 
  \end{matrix}
\end{array}\right]
\left[\begin{array}{@{}c@{}}
  \begin{matrix}
  q \\
  \dot{q}
  \end{matrix}
\end{array}\right] +
\left[\begin{array}{@{}c@{}}
  \begin{matrix}
  0 \\
  I
  \end{matrix}
\end{array}\right]u
\end{equation}
where 
\begin{equation}
\begin{cases}
u = -M^{-1}(\tau - g) = \left[ 0 \text{ } 0 \text{ } 0 \text{ } u_1 \text{ } u_2 \text{ } u_3 \right]^T \\
\left[ \dot{q} \text{ } \ddot{q} \right]^T = \left[ q_1 \text{ } q_2 \text{ } q_3 \text{ } \dot{q_1} \text{ } \dot{q_2} \text{ } \dot{q_3} \right]^T
\end{cases}
\end{equation}
During the operation, the desired joint angle $q_d$ is set as a constant. The error between desired joint angle and real joint angle is set as:
\begin{equation}
e = q_d - q \label{eq}
\end{equation}
Then, 
\begin{equation}
\dot{e} = \dot{q}_d - \dot{q} = -\dot{q} \label{eq}
\end{equation}
The state-space model can be rewritten as:
\begin{equation}
\left[\begin{array}{@{}c@{}}
  \begin{matrix}
  \dot{e} \\
  \ddot{q}
  \end{matrix}
\end{array}\right] = 
\left[\begin{array}{@{}cc@{}}
  \begin{matrix}
  0 \\
  0
  \end{matrix}
  & 
  \begin{matrix}
  -I \\
  -M^{-1}N 
  \end{matrix}
\end{array}\right]
\left[\begin{array}{@{}c@{}}
  \begin{matrix}
  e \\
  \dot{q}
  \end{matrix}
\end{array}\right] +
\left[\begin{array}{@{}c@{}}
  \begin{matrix}
  0 \\
  I
  \end{matrix}
\end{array}\right]u
\end{equation}

Follows the form of:
\begin{equation}
\dot{x} = Ax + Bu \label{eq}
\end{equation}
Considering the fixed point:
\begin{equation}
\begin{cases}
\dot{e} = 0 \Rightarrow \dot{q}_f = 0 \\
\ddot{q} = 0 \Rightarrow u = M^{-1}N \Rightarrow \tau = g
\end{cases}
\end{equation}
\\
It indicates that when the joint angular velocity is zero and the input torque is the gravity of the joint, the system will be stable, which corresponds to the actual situation.

\subsection{Discretisation}
The Forward Euler method is applied to do the discretisation of the state-space equation. The derivative of $x$ can be written as:
\begin{equation}
\dot{x} = \frac{x(k+1) - x(k)}{T} \label{eq}
\end{equation}
where $T$ is the sampling time. Substituting (28):
\begin{equation}
\frac{x(k+1) - x(k)}{T} = A x(k) + B u(k) \label{eq}
\end{equation}
Therefore,
\begin{equation}
x(k+1) = (TA + I) x(k) + TB u(k) \label{eq}
\end{equation}
The discrete time state-space can be written as follows:
\begin{equation}
x(k+1) = A_d x(k) + B_d u(k) \label{eq}
\end{equation}
where $A_d = TA + I$, $B_d = TB$.

\section{Model Predictive Control}

Model predictive control obtains optimal results by predicting how the system will perform over a certain future period. Set $N$ as the predictive horizon. To minimise a quadratic cost function at time $t = k$, it predicts the state value at time $t = k+1, t = k+2,...,t = k+N$ but only takes the obtained first state. By receding horizon control, optimised results are obtained at each sampling time instant.
Set output $y = x$, when at time $t = k$:
\begin{subequations}\label{eq:litdiff}
\begin{align}
X_k^T &= [x_k, x_{k+1}, x_{k+2}, \cdots, x_{k+N}]^T\label{eq} \\
U_k^T &= [u_k, u_{k+1}, u_{k+2}, \cdots, u_{k+N-1}]^T\label{eq}
\end{align}
\end{subequations}
where $X_k$ and $U_k$ represent the sequences of input and predicted state value respectively. The discrete time state-space can be written as:

\begin{equation}
\begin{split}
x_k &= x_k \\
x_{k+1} &= A_d x_k + B_d u_k \\
x_{k+2} &= A_d^2 x_k + A_d B_d u_k + B_d u_{k+1} \label{eq} \\
\vdots  \\
x_{k+N} &= A_d^N x_k + A_d^{N-1} B_d u_k + \cdots + B_d u_{k+N-1}
\end{split}
\end{equation}
This series of equations can be rewritten in the matrix form:
\begin{equation}
X_k = C X_k + F U_k\label{eq}
\end{equation}
\begin{equation*}
C = 
\left[\begin{array}{@{}c@{}}
  \begin{matrix}
  I \\
  A_d \\
  A_d^2 \\
  \vdots \\
  A_d^N
  \end{matrix}
\end{array}\right], \text{}
F = 
\left[\begin{array}{@{}cccc@{}}
  \begin{matrix}
  0 \\
  B_d \\
  A_dB_d \\
  \vdots \\
  A_d^{N-1}B_d
  \end{matrix}
  & 
  \begin{matrix}
  0 \\
  0 \\
  B_d \\
  \vdots \\
  A_d^{N-2}B_d
  \end{matrix}
  & 
  \begin{matrix}
  \cdots \\
  \cdots \\
   \\
  \ddots \\
  \cdots
  \end{matrix}
  & 
  \begin{matrix}
  0 \\
  0 \\
  0 \\
  \vdots \\
  B_d
  \end{matrix}
\end{array}\right]
\end{equation*}
As \cite{b10} proposed, the optimisation problem is to minimise the cost function:
\begin{equation}
J = X_{k+N}^T P X_{k+N} + \sum_{s=0}^{N-1}\left( X_{k+s}^T Q X_{k+s} + U_{k+s}^T R U_{k+s} \right) \label{eq}
\end{equation}
According to \cite{b10}, by substituting (36) into (37), the cost function can be rewritten in the quadratic form, which is also the function to be minimized:
\begin{equation}
J = \frac{1}{2}X_{k}^T Y X_{k} + \min\left( X_{k}^T M U_{k} + \frac{1}{2}U_{k}^T H U_{k} \right) \label{eq}
\end{equation}
where $Y$, $M$ and $U$ are obtained from $Q$, $R$, $P$, $C$ and $F$.

\section{Results of Simulation}
\vspace{-5pt}
This section will combine the previously mentioned techniques to demonstrate how model predictive control can be used to reduce operating errors.

Firstly, as shown in Fig.~\ref{fig7}, when the camera sees an object, it first identifies the object according to its colour, and then calculates the depth and three-dimensional coordinates of the object in the camera according to the comparison between the size of the object seen and the actual size. Then, the coordinate transformation and inverse kinematics are applied to get the angle that each joint needs to rotate to reach the target. For example, the recognised coordinates are $(10, 15, 10)$\,cm, and the desired joint angle is $(28.1, 67.8, 53.8)$\,degrees.

After model predictive control, the input torque required by the manipulator to reach this coordinate will be obtained. Due to the action of this torque, the error between the space coordinates of the end effector and the coordinates of the object is reduced with MPC. Finally, the robotic arm is able to grasp the target more accurately. Also, the tracking effect and energy consumption of the controller can be changed by adjusting the matrices $Q$, $R$ and $P$, which are more in line with the actual needs.

\begin{figure}[htbp]
\centering
\includegraphics[scale=0.34]{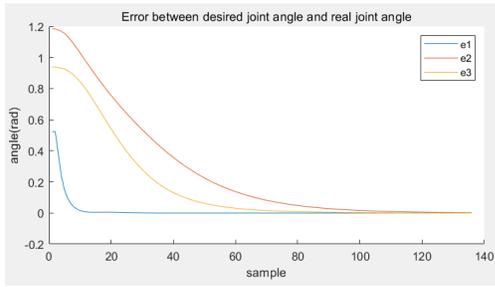}       
\caption{Error between desired joint angle and real joint angle.}
\label{fig9}
\end{figure}

\begin{figure}[htbp]
\centering
\includegraphics[scale=0.34]{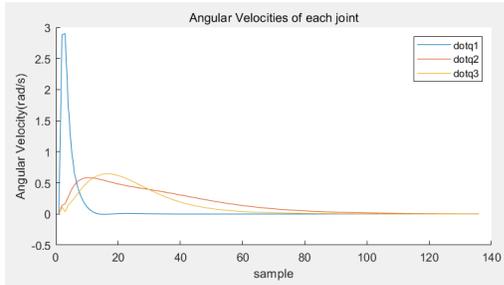}       
\caption{Angular Velocities of each joint.}
\label{fig10}
\end{figure}

\begin{figure}[htbp]
\centering
\includegraphics[scale=0.42]{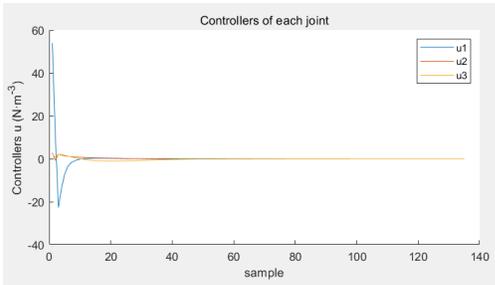}       
\caption{Controllers of each joint.}
\label{fig11}
\end{figure}

\begin{figure}[htbp]
\centering
\includegraphics[scale=0.34]{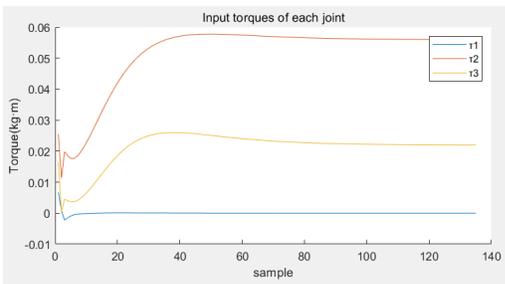}       
\caption{Input torques of each joint.}
\label{fig12}
\end{figure}

Set the sampling period as 0.05 seconds. The initial angle set of the three joints is $(0, 0, 0)$ . The desired angle of three joints is $(28.1\pi/180, 67.8\pi/180, 53.8\pi/180)$. According to the design of the state-space model \,(27), Fig.~\ref{fig9} represents the error between desired joint angle and real joint angle. The results of errors end up at zero indicating that each joint moves to the target position. Fig.~\ref{fig9},~\ref{fig10},~\ref{fig11} and~\ref{fig12} have been obtained by setting proper values of matrices $Q$, $R$ and $P$ to minimise the cost function. From Fig.~\ref{fig9},~\ref{fig10} and~\ref{fig12} we can see that, initially, the input torque of joint\,1 has provided a large angular velocity, enabling it to rotate to the target position first. Then, as joint\,2 and joint\,3 move closer to the target, their torques have increased and finally tend to a constant to support their gravity. The torque of chosen servo (DS3218) is $0.2$\,$kg \cdot m$, which is able to hold the whole system. The robotic arm tends to be stable in 6 seconds.

Fig. 13 and Fig. 14 show the initial and final position of the robotic arm, which is run by the co-simulation of software MATLAB and SolidWorks. This gesture of the robotic arm is selected because this is the coordinate where the torque demand of the servo is large. The simulated results illustrate that the torque of servos is capable of supporting the robotic arm and doing the grasping.

\begin{figure}[tbp]
\centering
\includegraphics[scale=1.0]{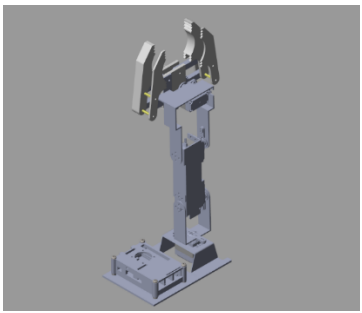}       
\caption{Initial position of the robotic arm.}
\label{fig13}
\end{figure}

\begin{figure}[htbp]
\centering
\includegraphics[scale=0.9]{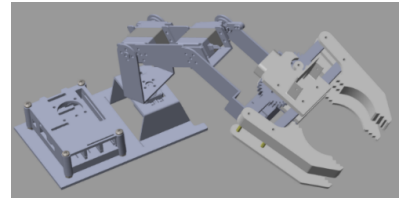}       
\caption{Final position of the robotic arm.}
\label{fig14}
\end{figure}

\vspace{18pt}

\section{Conclusion}
This project used a monocular camera to identify the colour and depth of the object, thus obtaining the spatial coordinates of the object through coordinate transformation. Through inverse kinematics calculation, the angle that each servo needs to rotate was obtained. Finally, model predictive control was applied to simulate and got the optimal input torques of servos, making the robotic arm reach the target point accurately and grasp the object. For recognition, it is convenient to identify objects of different colours by adjusting the HSV value. Also, this project can be applied to industrial scenarios such as sorting different packages in production. In the future, we will work on establishing the model more accurately and taking into account the disturbance of displacement and moment. Besides, our controller is designed as (22) here. The torque $\tau$ we required is not chosen directly chosen as controller. In the future, we will focus on the improvement of controller and external environment simulation.

\vspace{12pt}

\end{document}